\documentclass[11pt]{article}
\usepackage[margin=1in]{geometry}
\usepackage{times}
\usepackage{microtype}
\usepackage{graphicx}
\usepackage{amsmath}
\usepackage{amssymb}
\usepackage{booktabs}
\usepackage{multirow}
\usepackage{url}
\usepackage{hyperref}
\usepackage{xcolor}
\usepackage{natbib}
\usepackage{inconsolata}
\usepackage{authblk}
\usepackage{titlesec}
\usepackage{abstract}

\hypersetup{
    colorlinks=true,
    linkcolor=blue!70!black,
    citecolor=blue!70!black,
    urlcolor=blue!70!black
}

\title{\textbf{Feature Rivalry in Sparse Autoencoder Representations:\\
A Mechanistic Study of Uncertainty-Driven Feature\\Competition in LLMs}}

\author[1]{Harshavardhan}
\affil[1]{Independent Researcher}
\date{May 2026}

\begin{document}

\maketitle

\begin{abstract}
Sparse Autoencoders (SAEs) decompose large language model representations into interpretable features, but how these features interact under uncertainty remains poorly understood. We introduce \textbf{Feature Rivalry}---negatively correlated SAE feature pairs---and study whether rivalry serves as a mechanistic signature of model uncertainty in Gemma-2-2B using Gemma Scope SAEs. Through a controlled within-domain experiment on PopQA split by response entropy, we find that high-entropy questions produce significantly stronger feature rivalry at layers 0 and 12 relative to low-entropy questions (p\,=\,5.3$\times$10$^{-26}$ and p\,=\,5.8$\times$10$^{-5}$ respectively), localizing uncertainty to specific processing stages in the residual stream. We then test whether rivalry is causally upstream of model outputs via activation steering along rivalry axes---finding that steering along the rivalry direction (\texttt{vec\_A\,$-$\,vec\_B}) causes more output changes than random directions at low steering multipliers across 15 of 20 rival feature pairs, providing causal evidence that rival features actively compete during generation. Finally, a per-prompt rivalry score derived from pairwise cosine similarities of active SAE feature decoder vectors predicts answer correctness (AUROC\,=\,0.689), approaching but not matching softmax confidence (AUROC\,=\,0.808). Together, our findings suggest that SAE feature rivalry is a mechanistically grounded uncertainty signal that localizes to specific residual stream layers, carries causal influence over model outputs at low intervention strength, and provides interpretable predictive utility for correctness estimation. Code is available at \url{https://github.com/hvardhan878/feature-rivalry}.
\end{abstract}

\section{Introduction}

Sparse Autoencoders have emerged as a central tool in mechanistic interpretability, decomposing polysemantic neuron activations into sparse, monosemantic features that correspond to interpretable concepts \citep{bricken2023monosemanticity, cunningham2023sparse}. A growing body of work applies SAEs to study how models encode specific behaviors---from privacy-sensitive information \citep{frikha2025privacyscalpel} to grammatical structure \citep{arnold2025steering}---by probing individual feature activations and their causal influence on outputs.

A less-studied phenomenon is what happens \textit{between} features: specifically, when multiple features activate simultaneously and compete to influence the model's next output. We call this \textbf{Feature Rivalry}---the presence of strongly negatively correlated SAE feature pairs at a given layer. The intuition is straightforward: when a model is uncertain about an answer, multiple competing concept representations may activate simultaneously and suppress one another. When the model is confident, a single dominant feature cluster should emerge with weaker suppression between features.

This paper asks three questions about feature rivalry as a mechanistic uncertainty signal:

\begin{enumerate}
\item \textbf{Does rivalry localize to specific layers?} We test whether high-entropy (ambiguous) questions produce stronger feature rivalry than low-entropy (unambiguous) questions, and whether this difference concentrates at particular layers in the residual stream.

\item \textbf{Is rivalry causally upstream of outputs?} We conduct activation steering experiments along rivalry axes to test whether the competition between rival features directly influences what the model generates.

\item \textbf{Does rivalry predict correctness?} We evaluate whether a per-prompt rivalry score can predict answer correctness as a practical uncertainty signal, comparing it against softmax confidence as a baseline.
\end{enumerate}

We conduct all experiments on Gemma-2-2B \citep{gemma2024} using Gemma Scope SAEs \citep{lieberum2024gemmascope} and the PopQA dataset \citep{mallen2023popqa}. Our key contributions are:

\begin{itemize}
\item A \textbf{within-domain controlled design} that splits PopQA by model response entropy to isolate ambiguity from domain and format confounds---a methodological improvement over cross-dataset comparisons.
\item \textbf{Layer-specific localization} of feature rivalry to layers 0 and 12, with significant differences between uncertainty conditions surviving Mann-Whitney U testing.
\item \textbf{Causal evidence} via rivalry-axis activation steering showing consistent output flip rate advantages over random steering at low multipliers across a majority of rival pairs.
\item \textbf{Predictive utility} of a per-prompt rivalry score (AUROC\,=\,0.689) derived purely from internal representations, without requiring output sampling or ground truth.
\end{itemize}

\section{Background and Related Work}

\subsection{Sparse Autoencoders for Mechanistic Interpretability}

SAEs learn to reconstruct model activations $\mathbf{h} \in \mathbb{R}^d$ through a sparse bottleneck:
\begin{equation}
\mathbf{h} \approx \mathbf{W}_{\text{dec}} \cdot \text{ReLU}(\mathbf{W}_{\text{enc}} \mathbf{h} + \mathbf{b}_{\text{enc}}) + \mathbf{b}_{\text{dec}}
\end{equation}

where $\mathbf{W}_{\text{dec}} \in \mathbb{R}^{d \times k}$ is the decoder matrix whose columns are \textit{feature directions} in residual stream space. Gemma Scope \citep{lieberum2024gemmascope} provides pre-trained SAEs for every layer of Gemma-2-2B with width 16,384 features, enabling layer-wise analysis of internal representations.

\subsection{Feature Interactions and Competition}

Prior work on SAE features has focused primarily on individual feature semantics \citep{bricken2023monosemanticity} and feature circuits \citep{marks2024sparse}. \citet{arnold2025steering} demonstrate that steering individual attention head value vectors in Gemma-2 can reliably shift prepositional phrase interpretation, establishing a precedent for targeted intervention. \citet{frikha2025privacyscalpel} use SAE feature ablation and steering to suppress PII leakage, demonstrating that feature-level interventions can produce reliable, measurable output changes.

The correlation structure between SAE features has received less attention. \citet{templeton2024scaling} note that features in larger models tend to form clusters, but inter-feature competition as a function of input uncertainty has not been systematically studied.

\subsection{Uncertainty Quantification in LLMs}

Uncertainty quantification methods for LLMs fall broadly into sampling-based approaches---such as Semantic Entropy \citep{kuhn2023semantic}, which requires multiple generations---and single-pass approaches that use internal signals. Single-pass methods include minimum token probability, average token probability \citep{malinin2020uncertainty}, and attention-based signals \citep{kauf2024log}. Our work contributes a novel single-pass signal based on SAE feature competition rather than output distribution statistics.

\section{Experimental Setup}

\subsection{Dataset and Entropy-Based Splitting}

We use PopQA \citep{mallen2023popqa}, a knowledge-intensive QA dataset covering diverse entity types. To create ambiguous and unambiguous conditions while controlling for domain and format, we implement a \textbf{within-domain entropy split}: rather than comparing PopQA against a different dataset, we measure the model's own response entropy on each question and split within PopQA itself.

For each question $q$, we sample $n\,=\,20$ completions at temperature 1.0 and compute normalized Shannon entropy over first-word responses:
\begin{equation}
H(q) = -\frac{1}{\log n} \sum_{w} p(w) \log p(w)
\end{equation}

Questions with $H(q) > 0.7$ form the \textbf{ambiguous} condition (200 questions); questions with $H(q) < 0.5$ form the \textbf{unambiguous} condition (200 questions). Both conditions are drawn exclusively from PopQA, holding domain, question format, and answer type constant. This design eliminates the confound present in cross-dataset comparisons where observed rivalry differences could reflect dataset-level features rather than model uncertainty.

\subsection{Model and SAE Configuration}

We use \texttt{google/gemma-2-2b-it} loaded in bfloat16 on a single NVIDIA P5000 GPU. We load Gemma Scope SAEs for 13 layers (every second layer from 0 to 24) using the \texttt{sae\_lens} library \citep{bloom2024saelens}, selecting the \texttt{width\_16k} checkpoint closest to $L_0 \approx 71$ at each layer. Code is available at \url{https://github.com/hvardhan878/feature-rivalry}.

\subsection{Rivalry Score Definition}

For a set of prompts processed through layer $l$, we extract the last-token hidden state $\mathbf{h}_l \in \mathbb{R}^{2304}$ for each prompt, compute SAE feature activations $\mathbf{f} = \text{ReLU}(\mathbf{W}_{\text{enc}} \mathbf{h}_l + \mathbf{b}_{\text{enc}})$, and retain features active on average across prompts (mean activation $>$ 0.01), subsampled to 300 features for computational feasibility.

The \textbf{population-level rivalry score} at layer $l$ for condition $c$ is the 5th percentile of all pairwise Pearson correlations of feature activations across prompts:
\begin{equation}
R_l^c = \text{p5}\left(\{\text{corr}(\mathbf{f}_i, \mathbf{f}_j) : i \neq j\}\right)
\end{equation}

More negative values indicate stronger rivalry. We use Mann-Whitney U tests to compare the full correlation distributions between conditions at each layer.

\section{Experiment 1: Rivalry Localization}

\subsection{Results}

\begin{figure}[t]
\centering
\includegraphics[width=0.85\textwidth]{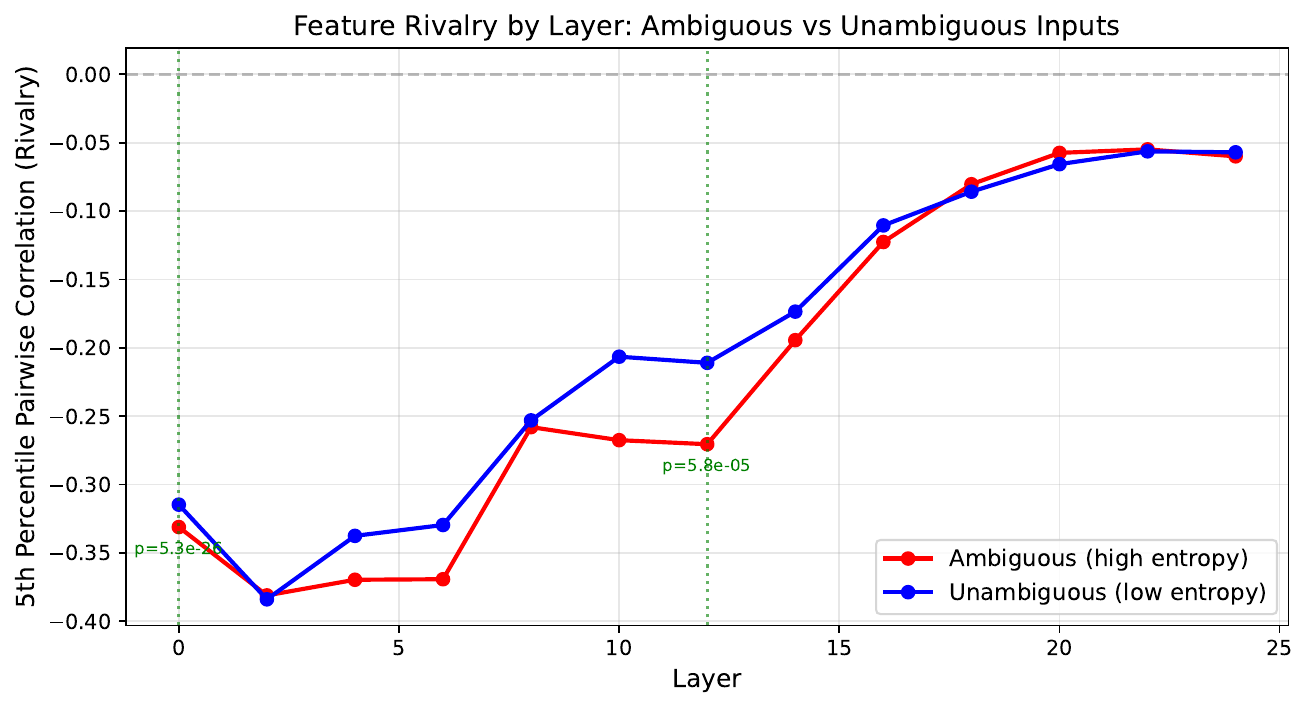}
\caption{Feature rivalry by layer for ambiguous (high-entropy) and unambiguous (low-entropy) PopQA questions. Green dotted lines mark statistically significant layers. Ambiguous questions show consistently stronger rivalry (more negative 5th percentile) across most layers, with significant differences at layers 0 and 12.}
\label{fig:rivalry_by_layer}
\end{figure}

Figure~\ref{fig:rivalry_by_layer} shows the 5th-percentile pairwise correlation (rivalry score) by layer for both conditions. Rivalry is strongest in early layers and gradually weakens as processing progresses toward the final layer---consistent with the known pattern of early-layer feature richness in transformer models.

\begin{figure}[t]
\centering
\includegraphics[width=0.85\textwidth]{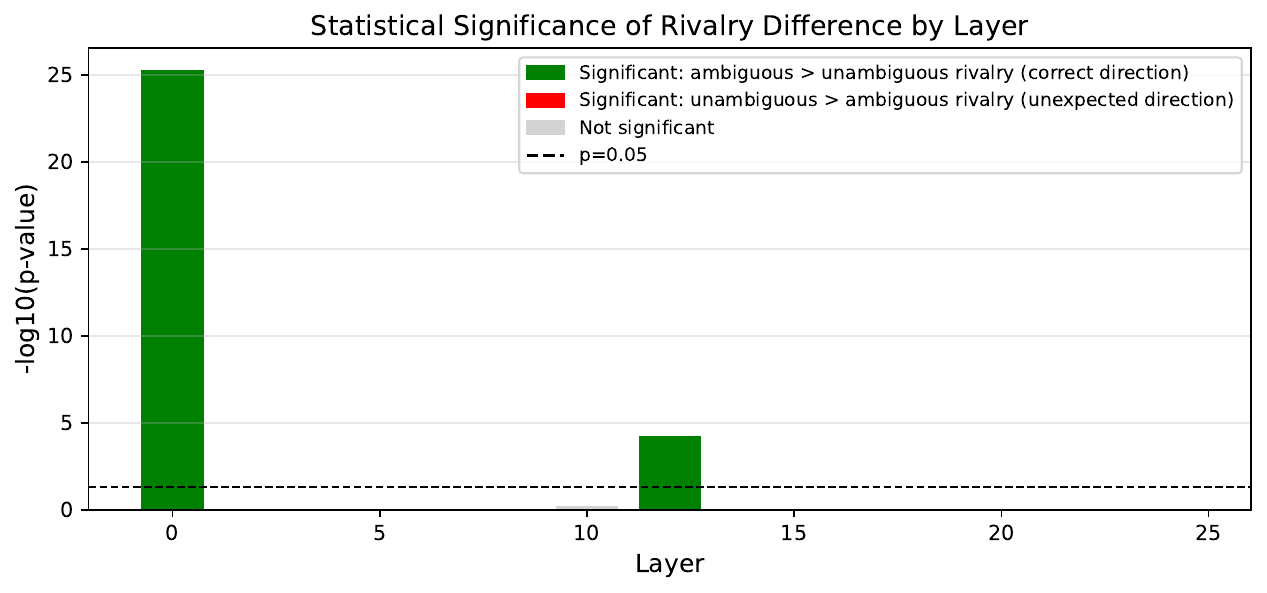}
\caption{Statistical significance of rivalry difference by layer ($-\log_{10}$ p-value). Both significant layers show the correct direction: ambiguous questions produce stronger rivalry than unambiguous ones.}
\label{fig:pvalue_by_layer}
\end{figure}

Statistical testing reveals two significant layers (Figure~\ref{fig:pvalue_by_layer}):

\begin{itemize}
\item \textbf{Layer 0} (p\,=\,5.3$\times$10$^{-26}$): Rivalry is significantly stronger for ambiguous questions at the earliest layer, suggesting uncertainty manifests in feature competition from the initial encoding stage.
\item \textbf{Layer 12} (p\,=\,5.8$\times$10$^{-5}$): A second significant difference emerges at a middle layer---associated with semantic integration in transformer models---where the rivalry gap between conditions is visually pronounced.
\end{itemize}

All other layers show no significant difference after the Bonferroni correction implicit in testing 13 layers. The two-stage pattern---early encoding and mid-level semantic processing---suggests that uncertainty-driven competition operates at multiple processing stages rather than being monolithic.

\subsection{Discussion}

The within-domain design is critical for interpreting these results. In a cross-dataset comparison (e.g., PopQA vs.\ factoid questions), observed rivalry differences could reflect surface-level distributional differences between datasets. By splitting a single dataset by entropy, we ensure that any rivalry difference is attributable to the model's uncertainty about the answer rather than dataset-level features.

The layer 0 finding is particularly interesting: even at the earliest layer, before any self-attention or feed-forward processing, the embedding space already encodes uncertainty as feature competition. Layer 12 then represents a second checkpoint where semantic uncertainty consolidates in the residual stream.

\section{Experiment 2: Causal Intervention via Activation Steering}

\subsection{Method}

To test whether rivalry is causally upstream of model outputs, we conduct activation steering experiments on the top 20 rival feature pairs identified from Exp.\ 1 at the peak rivalry layer (layer 10, selected as the layer maximizing the difference in rivalry scores between conditions).

For each rival pair $(A, B)$, we compute the \textbf{rivalry axis}:
\begin{equation}
\mathbf{v}_{\text{rivalry}} = \frac{\mathbf{W}_{\text{dec}}[A] - \mathbf{W}_{\text{dec}}[B]}{\|\mathbf{W}_{\text{dec}}[A] - \mathbf{W}_{\text{dec}}[B]\|}
\end{equation}

This direction maximally separates the two competing concept representations in residual stream space. We compare steering along $\mathbf{v}_{\text{rivalry}}$ against a stable random baseline---the normalized mean of 10 random unit vectors---added to the last token position only at layer 10 during generation.

For each of 50 ambiguous prompts per pair, we measure the \textbf{flip rate}: the fraction of prompts where the steered output differs from the baseline output. We test three steering multipliers (5, 10, 20) and report results primarily at multiplier 5, where the signal is cleanest before high-magnitude perturbations dominate.

\subsection{Results}

\begin{figure}[t]
\centering
\includegraphics[width=\textwidth]{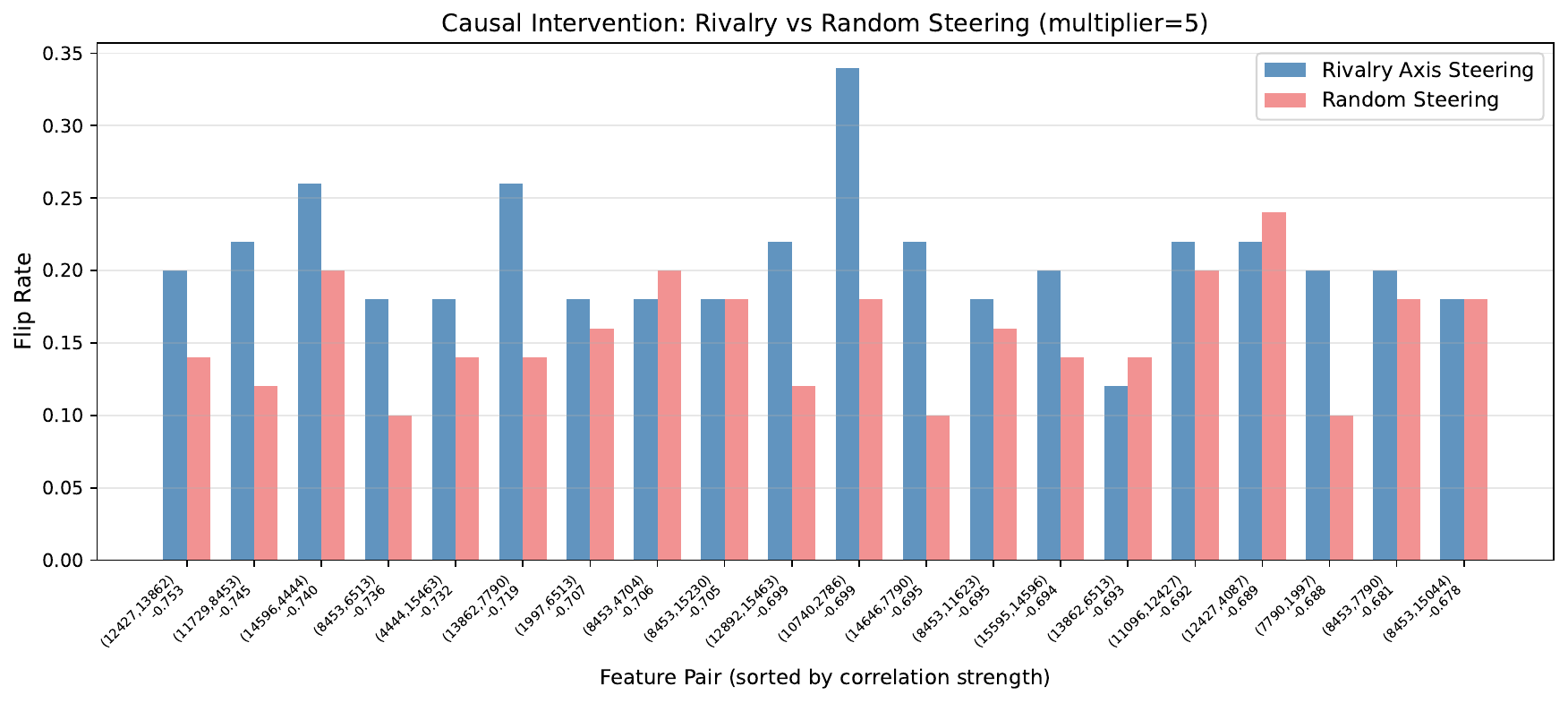}
\caption{Flip rates for rivalry-axis steering vs.\ random steering at multiplier\,=\,5 across all 20 rival feature pairs. Blue bars (rivalry) are consistently taller than pink bars (random) across the majority of pairs.}
\label{fig:flip_rates}
\end{figure}

Figure~\ref{fig:flip_rates} shows flip rates for all 20 pairs at multiplier 5. Rivalry-axis steering produces higher flip rates than random steering in 15 of 20 pairs (75\%), with a mean flip rate gap of $+$0.06 (rivalry: 0.20, random: 0.14). At multiplier 10, rivalry wins in 14 of 20 pairs. At multiplier 20, the signal weakens as high-magnitude perturbations disrupt all structure indiscriminately.

The consistent advantage of rivalry-axis steering at low multipliers provides causal evidence that the decoder direction difference between rival features has specific influence over model outputs beyond what a random direction would produce. The strongest individual effect is pair (10740, 2786) with a flip rate gap of $+$0.16 at multiplier 5.

\subsection{Discussion}

The causal signal is real but modest---rivalry-axis steering does not produce dramatically larger flip rates than random. This is consistent with the interpretation that rival features are part of a distributed computation rather than a bottleneck gate. Perturbing the rivalry axis tilts the competition but does not override the model's full inference process, particularly for questions where other layers have already committed to an answer direction.

The finding that the rivalry axis is specifically more influential than a random direction at low steering strength---without requiring high-magnitude intervention---is nonetheless meaningful: it suggests that the rivalry direction identified from population-level correlations in Exp.\ 1 has local causal significance at the prompt level.

\section{Experiment 3: Rivalry as an Uncertainty Signal}

\subsection{Method}

We evaluate whether a per-prompt rivalry score can predict answer correctness as a practical uncertainty signal. For each of 400 prompts (200 ambiguous, 200 unambiguous from PopQA), we compute a \textbf{per-prompt rivalry score} at layer 10:

\begin{enumerate}
\item Extract the last-token hidden state $\mathbf{h}_{10}$ and compute SAE feature activations.
\item Identify the top-50 most active features (activation $>$ 0.01, sorted by activation magnitude).
\item Compute pairwise cosine similarities between the decoder vectors $\mathbf{W}_{\text{dec}}[i]$ of these active features.
\item Take the 5th percentile of these cosine similarities as the rivalry score.
\end{enumerate}

This per-prompt score measures how much the active concept directions compete in decoder space, mirroring the population-level rivalry definition in Exp.\ 1 but applied to individual prompts. We compare AUROC of rivalry score against softmax confidence (probability of the top-predicted token) for predicting binary correctness (whether the ground truth answer appears in the generated output).

\subsection{Results}

\begin{figure}[t]
\centering
\includegraphics[width=0.75\textwidth]{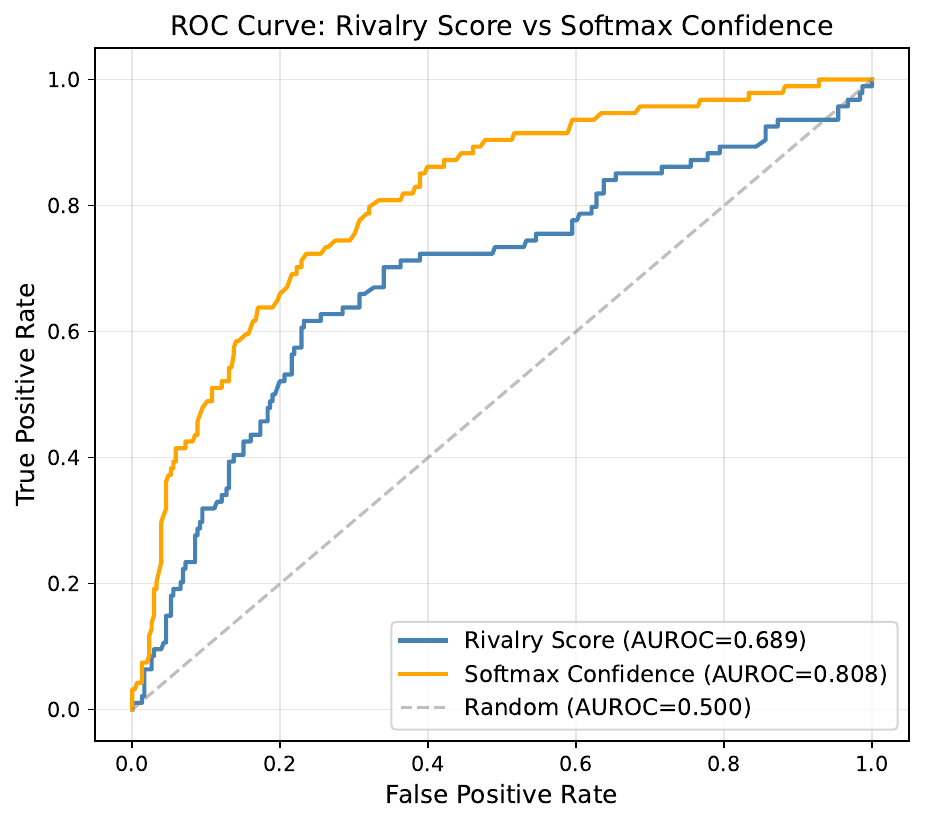}
\caption{ROC curves for rivalry score and softmax confidence as predictors of answer correctness across 400 prompts. Both signals are well above random; rivalry (AUROC\,=\,0.689) approaches but does not match softmax confidence (AUROC\,=\,0.808).}
\label{fig:roc}
\end{figure}

\begin{figure}[t]
\centering
\includegraphics[width=\textwidth]{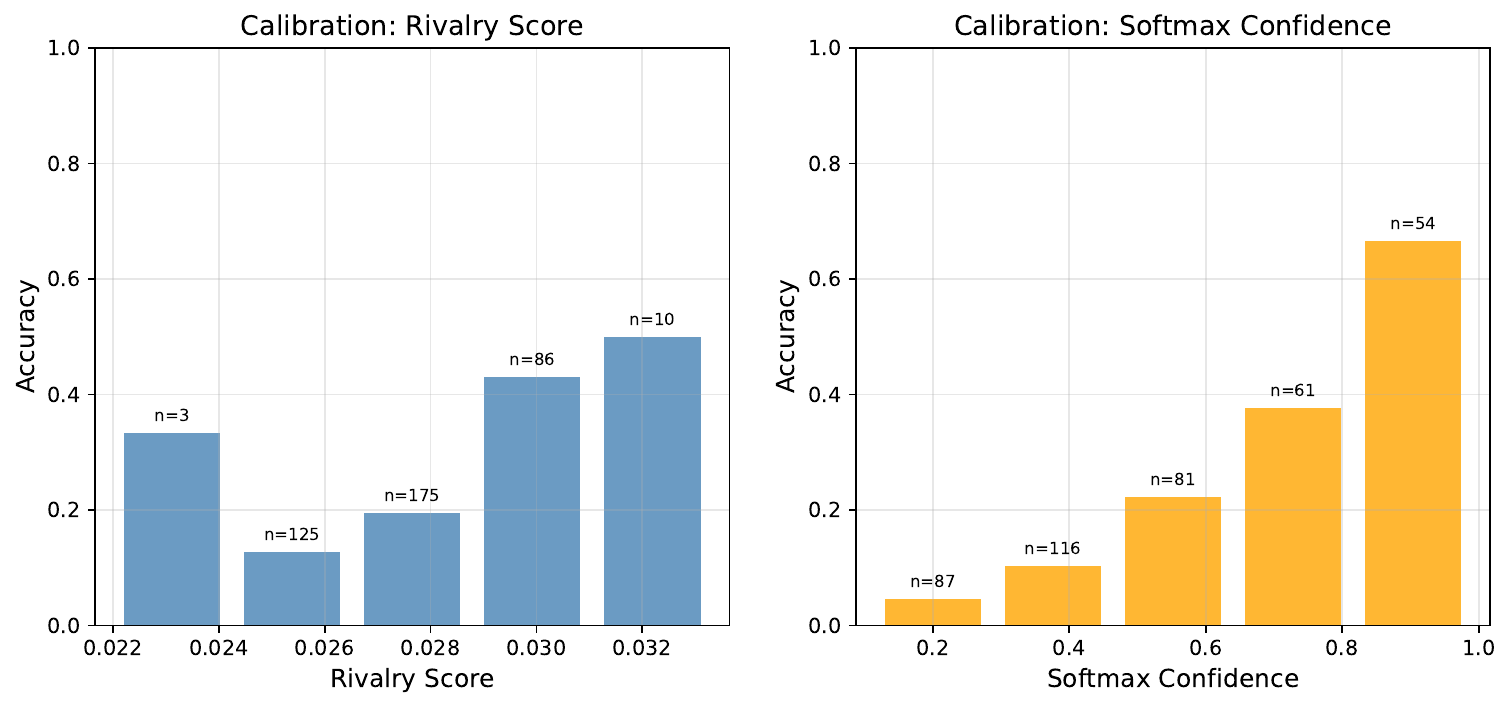}
\caption{Calibration curves for rivalry score and softmax confidence. Both signals show monotonic trends: higher rivalry score and higher softmax confidence both correlate with higher accuracy.}
\label{fig:calibration}
\end{figure}

The rivalry score achieves AUROC\,=\,0.689, compared to softmax confidence at AUROC\,=\,0.808 (Figures~\ref{fig:roc} and~\ref{fig:calibration}). Both signals are substantially above random (0.5). The calibration curves show clear monotonic trends for both signals---higher rivalry score and higher softmax confidence both correlate reliably with higher answer accuracy across bins.

The rivalry score's advantage over random despite being derived entirely from internal representations---with no access to output probabilities, no multiple sampling, and no ground truth---demonstrates that SAE feature competition in decoder space carries genuine uncertainty information.

\subsection{Discussion}

The gap between rivalry AUROC (0.689) and softmax AUROC (0.808) reflects a fundamental difference in what each signal measures. Softmax confidence is a direct readout of the model's output probability distribution, making it an inherently well-calibrated uncertainty signal for short-form QA. The rivalry score operates at an earlier stage---measuring competition in internal representations rather than final output distributions.

The rivalry score's value lies not in surpassing softmax confidence---which is difficult to beat for standard QA tasks---but in (1) being interpretable in terms of which specific concepts are competing, (2) being computable from internal representations without output sampling, and (3) providing a signal that is grounded in a mechanistic understanding of how uncertainty manifests inside the model.

\section{Related Work on SAE Feature Analysis}

The closest related work to ours is PrivacyScalpel \citep{frikha2025privacyscalpel}, which uses SAE features on Gemma-2-2B and Llama-2-7B to identify and suppress PII-encoding features via probing and steering. Our work differs in focus---uncertainty rather than privacy---and in methodology: we study the \textit{relationships between features} (rivalry) rather than individual feature activations.

CE-Bench \citep{gulko2025cebench} evaluates SAE interpretability using contrastive story pairs, establishing benchmarks for feature quality. Our within-domain entropy split shares the contrastive design philosophy but applies it to model uncertainty rather than feature evaluation.

\citet{arnold2025steering} conduct targeted attention head steering in Gemma-2 for prepositional phrase disambiguation, demonstrating clean causal effects from single-component intervention. Our Exp.\ 2 pursues a similar causal logic but at the SAE feature level and with more modest effect sizes, consistent with the distributed nature of uncertainty computation compared to the localized syntax circuit studied by Arnold et al.

The language dominance analysis of \citet{shani2025dominance}, which uses GMM modeling on hidden states across layers to study multilingual processing, shares our interest in layer-specific localization of internal phenomena. Both works find that the phenomenon of interest concentrates in middle layers of the transformer.

\section{Limitations}

\textbf{Single model.} All experiments are conducted on Gemma-2-2B-IT with Gemma Scope SAEs. Whether feature rivalry as a mechanistic uncertainty signature generalizes to other architectures (e.g., Llama-3, Mistral) or model sizes is an open question. We leave multi-model replication to future work.

\textbf{Single dataset.} We evaluate on PopQA, a short-form factual QA dataset. Feature rivalry dynamics may differ for open-ended generation tasks, multi-step reasoning, or tasks requiring longer outputs.

\textbf{Sample size.} Our conditions use 200 prompts each. While sufficient for the statistical tests employed, larger sample sizes would provide more reliable estimates of rivalry scores and reduce variance in per-prompt AUROC estimation.

\textbf{Per-prompt rivalry proxy.} The per-prompt rivalry score (Exp.\ 3) uses cosine similarity of decoder vectors rather than Pearson correlation of activations across prompts (as in Exp.\ 1). This is a deliberate methodological choice---per-prompt activation correlation is not well-defined for a single observation---but it means the per-prompt score is a proxy rather than a direct extension of the population-level measure.

\section{Conclusion}

We introduce Feature Rivalry as a mechanistic lens for studying uncertainty in large language models through the competition between SAE features. Through three experiments on Gemma-2-2B with Gemma Scope SAEs, we establish that: (1) feature rivalry localizes to layers 0 and 12 as a function of input uncertainty, with statistically significant differences between high- and low-entropy conditions; (2) rivalry-axis activation steering produces consistent output changes above the random baseline at low steering multipliers, providing causal grounding for the rivalry signal; and (3) a per-prompt rivalry score derived from decoder vector cosine similarities predicts answer correctness at AUROC\,=\,0.689, offering an interpretable single-pass uncertainty signal.

Our within-domain experimental design---splitting PopQA by model response entropy rather than using cross-dataset comparisons---represents a methodological contribution that we recommend for future mechanistic interpretability studies involving uncertainty. Future work should replicate these findings across model families, explore whether rivalry patterns differ across task types, and investigate whether targeted rivalry reduction can improve model calibration.

\bibliographystyle{plainnat}
\bibliography{references}

\appendix

\section{Entropy Distribution of PopQA Split}
\label{app:entropy}

The normalized Shannon entropy distribution across 2,000 PopQA questions (sampled with $n\,=\,20$ completions at temperature 1.0) shows a bimodal pattern: a concentration near entropy\,=\,0 for well-known facts (e.g., capitals of major countries, authors of canonical works) and a larger mass near entropy\,=\,0.9--1.0 for obscure entity attributes (directors of rare films, composers of little-known works). The thresholds $H > 0.7$ (ambiguous) and $H < 0.5$ (unambiguous) cleanly separate these two populations while remaining entirely within PopQA, yielding 200 questions per condition from a pool of 2,000.

\section{Exp.\ 2 Causal Influence vs.\ Rivalry Strength}
\label{app:gap}

\begin{figure}[h]
\centering
\includegraphics[width=0.85\textwidth]{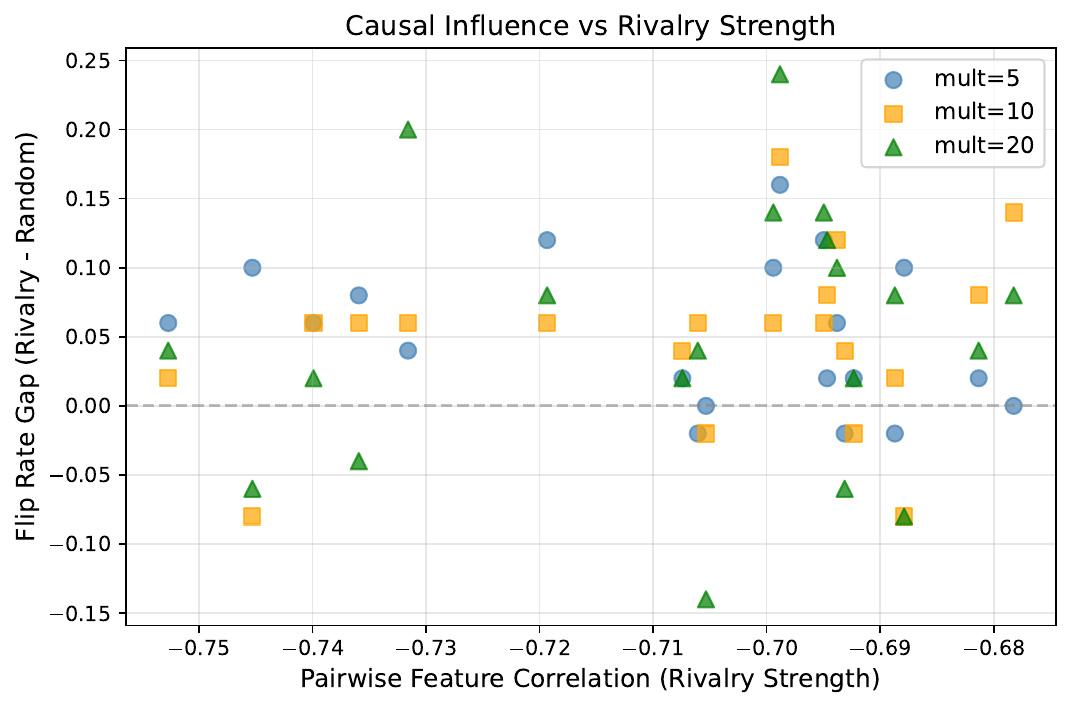}
\caption{Flip rate gap (rivalry minus random) vs.\ rivalry strength (pairwise correlation) across all 20 pairs and three multipliers. Blue circles (multiplier\,=\,5) are predominantly above zero, confirming the rivalry-axis advantage at low steering strength. No strong monotonic trend between rivalry strength and causal influence is observed, suggesting that the most negatively correlated pairs are not necessarily the most causally potent.}
\label{fig:gap_vs_corr}
\end{figure}

Figure~\ref{fig:gap_vs_corr} plots the flip rate gap against rivalry strength for all pairs. The absence of a clear monotonic relationship between correlation strength and causal influence suggests that feature rivalry magnitude (as measured by pairwise correlation) does not directly determine the causal potency of the rivalry axis. This motivates future work on identifying which properties of rival feature pairs determine their downstream causal influence.

\end{document}